\documentclass{article}

\usepackage{microtype}
\usepackage{enumitem}
\usepackage{graphicx}
\usepackage{subfigure}
\usepackage{booktabs} 

\usepackage{hyperref}

\usepackage[accepted]{icml2025}

\usepackage{amsmath}
\usepackage{amssymb}
\usepackage{mathtools}
\usepackage{amsthm}
\usepackage{xcolor}
\usepackage{colortbl}
\usepackage{float}

\usepackage[capitalize,noabbrev]{cleveref}

\theoremstyle{plain}

\theoremstyle{definition}

\theoremstyle{remark}

\usepackage[textsize=tiny]{todonotes}

\icmltitlerunning{Solo Connection}

\begin{document}

\twocolumn[
\icmltitle{Solo Connection: A Parameter Efficient \\ Fine-Tuning Technique for Transformers}



\icmlsetsymbol{equal}{*}

\begin{icmlauthorlist}
\icmlauthor{Harsh Nilesh Pathak}{WPI}
\icmlauthor{Randy Paffenroth}{WPI2}
\end{icmlauthorlist}

\icmlaffiliation{WPI}{Department of Data Science, Worcester Polytechnic Institute, MA, United States}
\icmlaffiliation{WPI2}{Department of Data Science, Mathematics and Computer Science, Worcester Polytechnic Institute, MA, United States}

\icmlcorrespondingauthor{Harsh Nilesh Pathak}{hnpathak@wpi.edu}
\icmlcorrespondingauthor{Randy Paffenroth}{rcpaffenroth@wpi.edu}

\icmlkeywords{Machine Learning, ICML}

\vskip 0.3in
]



\printAffiliationsAndNotice{\icmlEqualContribution} 

\begin{abstract}
Parameter-efficient fine-tuning (PEFT) is a versatile and extensible approach for adapting a Large Language Model (LLM) for newer tasks.  One of the most prominent PEFT approaches, Low-Rank Adaptation (LoRA), primarily focuses on adjusting the attention weight matrices within individual decoder blocks of a Generative Pre-trained Transformer (GPT-2). In contrast, we introduce Solo Connection—a novel method that adapts the representation at the decoder-block level rather than modifying individual weight matrices. Not only does Solo Connection outperform LoRA on E2E natural language generation benchmarks, but it also reduces the number of trainable parameters by 59\% relative to LoRA and by more than 99\% compared to full fine-tuning of GPT-2 one of the earliest versions of large language models (LLMs). Another key motivation for Solo Connection comes from homotopy theory, where we introduce a trainable linear transformation that gradually interpolates between a zero vector and the task-specific representation, enabling smooth and stable adaptation over time.

While skip-connections in the original 12-layer GPT-2 are typically confined to individual decoder blocks, subsequent GPT-2 variants scale up to 48 layers, and even larger language models can include 128 or more decoder blocks. These expanded architectures underscore the need to revisit how skip connections are employed during fine-tuning. This paper focuses on "long skip connections" that link outputs of different decoder blocks, potentially enhancing the model's ability to adapt to new tasks while leveraging pre-trained knowledge.
\end{abstract}

\section{Introduction}

\begin{table*}[h]
\centering
\begin{tabular}{|p{6cm}|c|c|c|c|c|c|}
\hline
Model & \#Params & BLEU & NIST & E2E MET & Rouge & CIDEr \\
\hline
FT GPT-2 M & 354.92M & 68.2 & 8.62 & 46.2 & 71.0 & 2.47 \\
\hline
Baseline LoRA GPT-2 Medium & 0.35M & 67.45 & 8.58 & 45.93 & 68.8 & 2.36 \\
Solo Connection GPT-2 Medium  & 0.26M & \cellcolor{green!80} 67.7 & \cellcolor{green!80} 8.64 & \cellcolor{green!80}45.95 & \cellcolor{green!80} 69.13 & \cellcolor{green!80}2.36\\

\hline
Baseline LoRA  GPT-2 Small & 0.29M & 65.79 & 8.49 & 45.21 & 67.46 & 2.27 \\
Solo Connection GPT-2 S  & 0.12M & \cellcolor{green!80} 67.64 & \cellcolor{green!80}8.64 & \cellcolor{green!80} 45.70 & \cellcolor{green!80} 68.32 & \cellcolor{green!80} 2.31 \\
\hline
\end{tabular}
\caption{Performance comparison of GPT-2 Medium (GPT-2 M) and GPT-2 Small (GPT-2 S) models using three approaches: fully fine-tuned (FT), LoRA, and our proposed Solo Connection. While the full fine-tuning of GPT-2 M requires 354.92M parameters, Solo Connection uses only 0.26M but still achieves comparable or superior results across NLG metrics compared to other methods. Solo Connection has 99\% less trainable parameters, demonstrating its efficiency and effectiveness in reducing parameter counts without sacrificing performance.} 
\label{tab:gpt1}
\end{table*}

\begin{figure*}
    \begin{center}
    \includegraphics[scale=0.23]{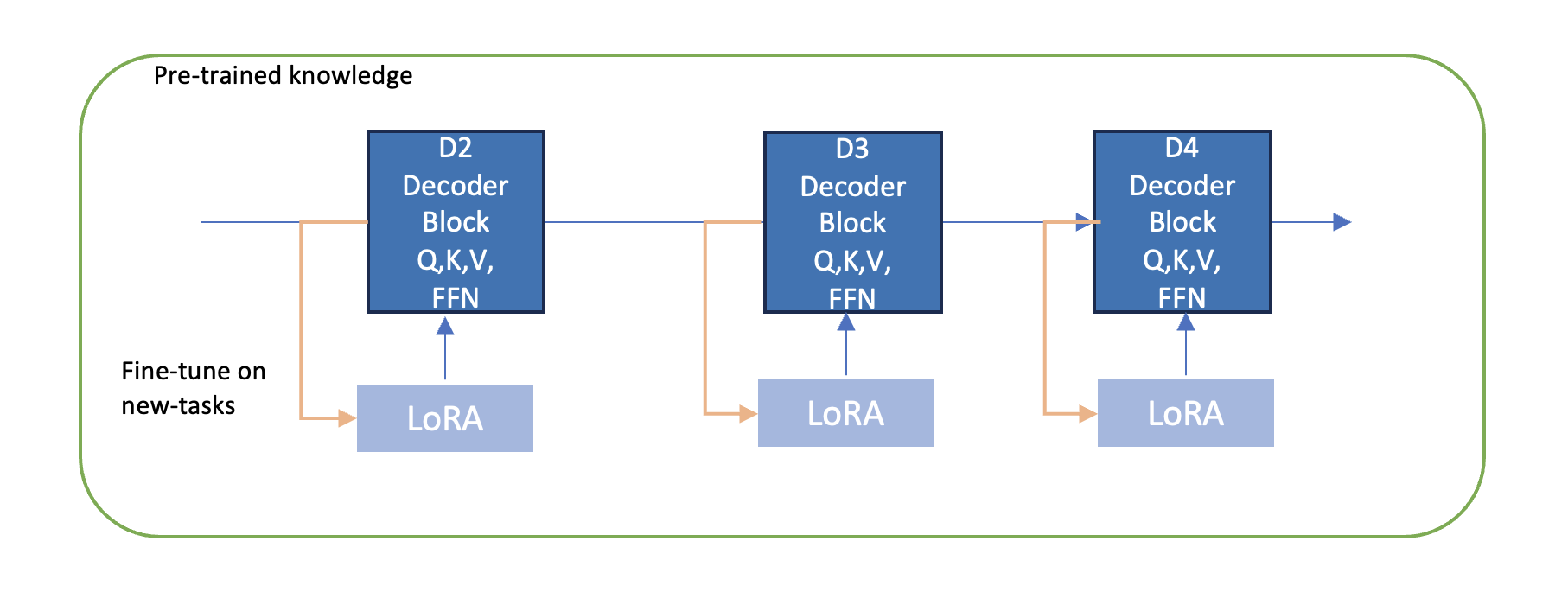}
    \includegraphics[scale=0.23]{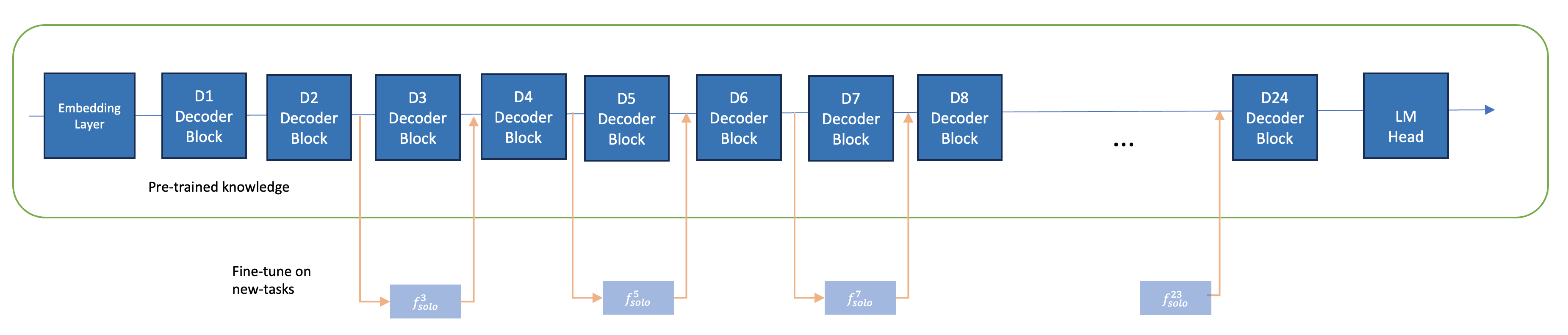}
    \caption{Solo Connection compared with LoRA setup}
    \end{center}
    \small{LoRA (top) is Intra-decoder connection: Fine-tuning large language models with LoRA involves adding trainable modules within each decoder layer, creating an intra-decoder connection. Solo Connection (bottom), on the other hand, introduces an inter-decoder connection where the shared encoder and decoder are connected across different decoder layers. This approach directly adapts the representation of the decoder blocks and we explore its potential in learning newer tasks. 
    }
    \label{fig:solor}
\end{figure*}

Pre-trained Language Models (PLMs) like GPT-2 \cite{radford2019gpt2}, GPT-3, GPT-4, LLAMA-2 \cite{touvron2023llama2openfoundation}, and Transformer-XL \cite{txl} have transformed NLP by leveraging self-supervised objectives such as language modeling. These models predict the next token given a sequence, enabling them to generate coherent text. Despite their success, adapting large PLMs to new domains remains resource-intensive, limiting accessibility for research groups with constrained compute and memory.

Parameter-efficient fine-tuning (PEFT) tackles this challenge by adapting PLMs using a small subset of parameters \cite{xu2023parameterefficient}. Techniques like LoRA \cite{hu2021lora}, BitFit \cite{ben-zaken-etal-2022-bitfit}, and Adapters \cite{adapter} have shown that models can retain strong performance without full fine-tuning. These methods are effective in low-resource settings and have been used in NLP, vision, and multi-modal tasks \cite{lecun2015deep, peft, pathakprincipled_harsh}.

We propose \textit{Solo Connection}, a sparse and low-rank fine-tuning strategy inspired by LoRA. Instead of modifying weights directly, it leverages long skip connections within decoder blocks to adapt model representations efficiently. Our approach emphasizes parameter sharing and sparsity, reducing the number of trainable parameters while preserving task performance. The core motivation behind \textit{Solo Connection} is to fundamentally shift the paradigm of parameter-efficient fine-tuning (PEFT) from \textit{intra-layer adaptation} to \textit{inter-layer adaptation}. While it may be loosely described using the language of adapters, such a view oversimplifies its theoretical foundation and architectural design. Unlike most adapter-based approaches—including LoRA and its variants—which focus on inserting adaptation modules within specific subcomponents like attention or feedforward layers, Solo Connection operates across layers. It introduces weighted skip connections that span multiple decoder blocks, allowing information to flow more effectively and enabling parameter sharing across a broader context. This cross-layer mechanism supports richer representation learning while significantly reducing the number of trainable parameters. Another key motivation for Solo Connection stems from homotopy theory, rooted in continuation methods \cite{Pathak_rethink, cont_method_Allgower, pathakprincipled_harsh, pathak2018npacs } and dynamical systems \cite{cont_method_Allgower, strogatz2001nonlinear}. To enable gradual and stable adaptation, we introduce a homotopy-inspired \cite{pathak2018parameter, pathak2019parameter} linear transformation that interpolates between a zero vector and the task-specific representation. This mechanism is governed by trainable parameters that control the extent of adaptation over time. Unlike abrupt modifications to network weights, this smooth transition facilitates progressive learning and inherently stabilizes training. It also acts as an implicit gating mechanism, scaling the Solo Connection output dynamically without manual tuning, offering a principled alternative to heuristic scaling used in methods like LoRA. The key contributions of the paper are as follows:
\begin{enumerate}[nosep]

    \item We introduce a novel PEFT method using weighted long skip connections across decoder blocks to reduce redundancy.
    \item Solo Connection adapts representations more effectively, achieving better performance with fewer parameters.
    \item On the E2E benchmark, our method outperforms LoRA and full fine-tuning while using 59\% fewer parameters.
    \item We analyze individual design components of Solo Connection in Appendix-\ref{a:exp} to explain its performance gains.
\end{enumerate}

We evaluate Solo Connection on GPT-2 across five natural language generation (NLG) tasks \cite{novikova-e2e}, showing that it consistently outperforms LoRA with up to 59\% fewer parameters. While our experiments use transformer-based models, Solo Connection is architecture-agnostic and applicable across domains.

\section{Methodology}\label{method}
\begin{figure*}
    \centering
    \includegraphics[scale=0.26]{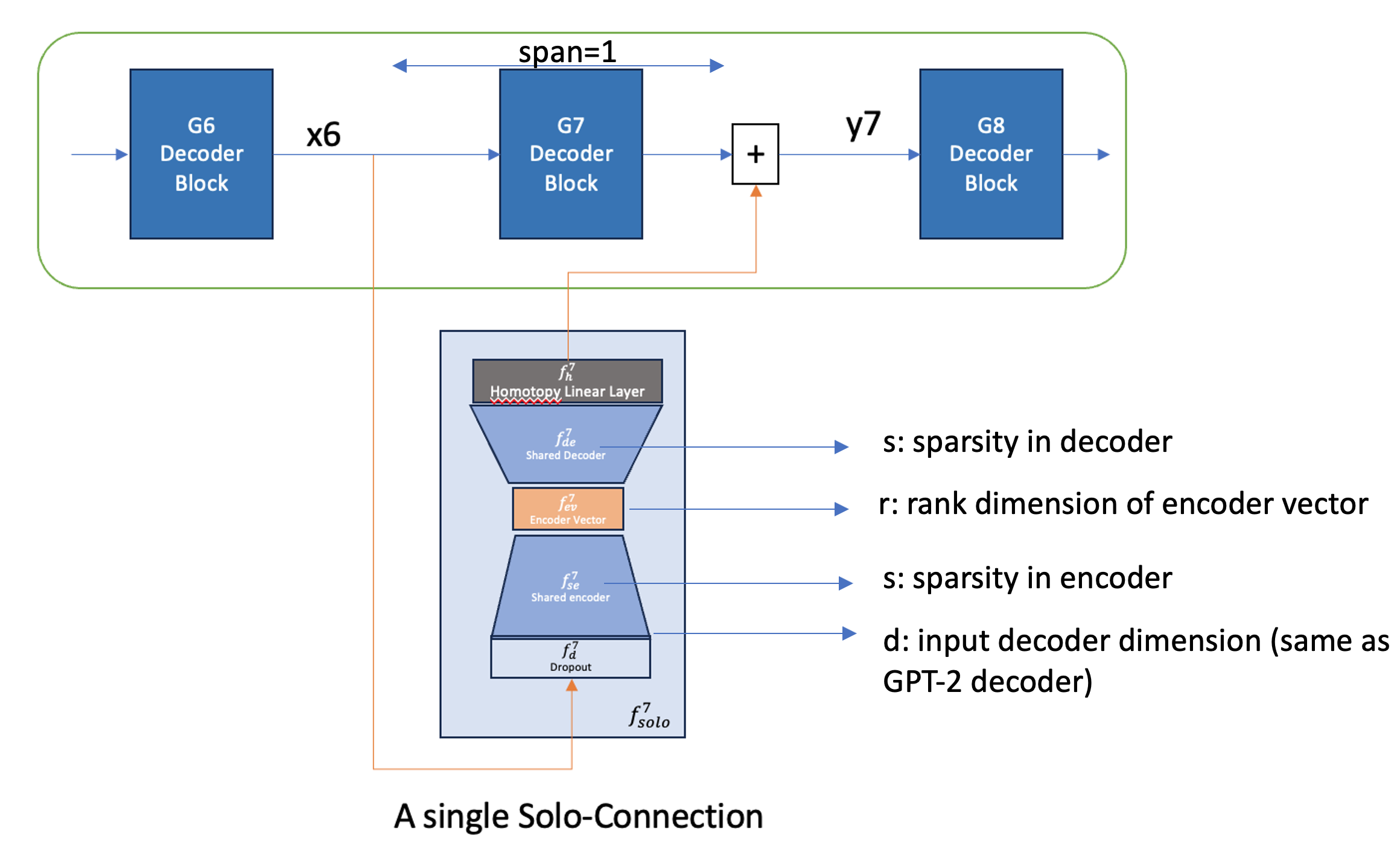}
    \caption{Solo Connection that transforms the representation from a previous decoder and adapts during fine-tuning for subsequent decoders. Here representation embedding of D6 is fine-tuned for $D_8$. Here we also show the building block of the solo connections, which has encoder-decoder learnable weights along with a homotopy layer.}
    \label{fig:solo}
\end{figure*}

Fine-tuning in \cite{lecun2015deep} is an essential technique in Deep learning to utilize the pre-trained knowledge for a downstream task. Solo Connection provides a unique perspective over well-adopted PEFT methods such as Adapters \cite{adapter}, LoRA \cite{hu2021lora} and its derivatives \cite{xu2023parameterefficient, adalora, vera}. Our method aims to adapt one or more pre-determined decoder blocks. Intuitively,  Solo Connection provides task-specific decoder representations for downstream tasks. The principles outlined here apply to any dense layers in deep learning models, though we only focus on GPT-2 language models in our experiments as the motivating use case.  Specifically, for the case of GPT-2 \cite{radford2019gpt2}, we observed that most LoRA-based techniques were applied to a weight-matrix; for example, it is used in the Query $Q$, Key $K$, or Value $V$ weight matrix. \emph{On the other hand, our method attempts to adapt the decoder block representation directly.} 

\subsection{The Solo Connection: Sparse and Low-rank Skip Connections}
We introduce Sparse and Low-rank Skip Connection (Solo Connection) as a trainable block that can be applied to fine-tune Pre-trained Language Models (PLMs). In GPT-2, a Solo Connection is added to select decoder blocks (e.g., $D_7$ in Figure-\ref{fig:solo}). It transforms the input ($x_6$) from a previous decoder block ($D_6$) before passing it to a subsequent block ($D_8$).
Equation-\ref{eq:solo2_gpt} defines the Solo Connection:
\begin{equation}\label{eq:solo2_gpt}
y_i = D_{i}(x_{i-1}, { \theta_{i}}) + f^{i}_\text{solo}(x_{i-1}, { \phi_{i}})
\end{equation}
Here, $D_{i}(x_{i-1}, { \theta_{i}})$ represents the \emph{fixed} pre-trained knowledge, while $f^{i}_\text{solo}(x_{i-1}, { \phi_{i}})$ is the \emph{adaptable} component for downstream tasks. ${\theta_i}$ and ${\phi_i}$ are the pre-trained and Solo Connection parameters, respectively. $x_{i-1} \in \mathbb{R}^d$ is the input from the previous decoder block, and $y_{i} \in \mathbb{R}^d$ is the output for the next block, where $d$ is the representation dimension (1024 for GPT-2 M, 768 for GPT-2 S).
Figure-\ref{fig:solo} details a single Solo Connection. By applying Solo Connections to alternate decoder blocks (Figure-\ref{fig:solor}), we aim to optimize downstream task performance. We test this hypothesis in Section-\ref{sec:experiment}.


We apply Solo Connections to alternate decoder blocks in GPT-2, starting from $D_2$ to the final block. GPT-2 Medium (24 decoders) and GPT-2 Small (12 decoders) results in 11 and 5 Solo Connections, respectively. Figure-\ref{fig:solor} illustrates this configuration, while Figure-\ref{fig:solo} details a single Solo Connection as implemented in our experiments. This approach aims to balance adaptability and efficiency in fine-tuning.

\subsection{The Building Block of Solo Connection}\label{csolo} 

This section will describe the components of ($f_\text{solo}$) Solo Connection. 

Equation-\ref{eq:solod} defines the composition of $f_\text{solo}$:
\begin{equation}\label{eq:solod}
    f_\text{solo} = f_{h} \circ f_\text{sd} \circ f_\text{ev} \circ f_\text{se} \circ f_{d}
\end{equation}

\noindent where, $f_{h}$ is the homotopy linear layer, $f_\text{sd}$ is the shared decoder, $f_\text{ev}$ is the encoding-vector, $f_\text{se}$ is the shared  encoder and $f_{d}$ is the dropout layer.

The shared encoder ($f_\text{se}$) transforms the input vector $\textbf{x}$ into a lower-dimensional representation of size $r$. This encoder is shared across all Solo Connection modules in the model (Figure~\ref{fig:solor}). The same $f_\text{se}$ function is applied to the inputs of each Solo Connection, regardless of its position in the network. For example, there are 5 Solo Connections in GPT-2 (S), then all 5 have a single trainable encoder $f_\text{se}$.  We also add a non-trainable dropout layer $f_d$ before the encoder to help generalize and improve metric performance. The shared Encoder incorporates two key hyperparameters. The first is dimensionality reduction by rank (r), which transforms the input to a lower-dimensional space of rank (r). Second, sparsity (s) - A hyperparameter $s$ randomly masks a fraction of parameters, setting $(s\%)$ to zero during both forward and backward passes.

The shared Decoder ($f_\text{sd}$) transforms the lower-dimensional representation back to the original input dimension. Like the shared encoder, the shared decoder is available across all Solo Connection modules.  The dimensionality reduction factor $r$ and sparsity hyperparameter $s$ are crucial for balancing efficiency and performance. Their impact and selection strategies are discussed in Section~\ref{sec:experiment}. We use Kaiming initialization \cite{HeZR015} for both the encoder and decoder functions in our method. This ensures the values are scaled based on the matrix dimensions, resulting in a consistent variance for all ranks when multiplying respective matrices. As a result, there is no need to fine-tune the learning rate for each rank \citet{vera, hu2021lora}. The shared encoder ($f_\text{se}$) and decoder ($f_\text{sd}$) are trainable, allowing for optimal encoding and decoding transformations.  The encoding vector ($f_\text{ev}$) is a trainable task-specific bias vector, while the Homotopy Linear layer ($f_{h}$) is a trainable linear transformation for output refinement. The dropout layer ($f_{d}$), though not trainable itself, aids in preventing overfitting during the training of other components.

Finally, we employ a Homotopy linear layer to learn from projection (\textbf{z}: output of $f_\text{sd}$) and gradually adapt to the new task.  This homotopy linear layer ($f_h$)  is a topological transformation between zero vector and the adapted representation $z \in \mathbb{R^d}$, as shown in Equation-\ref{eq:homotopy}.   

\begin{equation}\label{eq:homotopy}
    f_h(\textbf{z}) = \lambda \textbf{v} \odot \textbf{z} + (1-\lambda)   \textbf{0} 
\end{equation}

Here, $\lambda$ (scalar) and $\textbf{v} $ (vector) are trainable parameters, while $\textbf{z}$ is the output of the shared decoder. Also, value of $\lambda$ is bounded $[0,1]$. The homotopy layer is the Solo connection's dynamic scaling and gating mechanism.  As $\lambda$ increases from 0 to 1, the Solo connection gradually adapts to new tasks by incorporating more of $\textbf{z}$. Simultaneously, $\lambda$ acts as an automatic scalar for the Solo connection's output, eliminating the need for manual scaling as required in methods like LoRA. The homotopy layer is initially set to 0.001, allowing for gentle, gradual adaptation during fine-tuning.

\section{Experiments}\label{sec:experiment}

We evaluate our fine-tuning approach by replicating the experimental setup of LoRA \citep{hu2021lora}. All code and configurations are available on GitHub [Anonymous Link]. We use the E2E NLG Challenge dataset \citep{novikova-e2e}, a benchmark with diverse NLG tasks—making it ideal for testing generalization. This diversity ensures models trained here are well-suited for transfer across domains. Also, we fine-tune GPT-2 Small and Medium \citep{radford2019gpt2} using our proposed \textit{Solo Connection} and compare against full fine-tuning and LoRA baselines. GPT-2 models are widely used in both NLP and vision tasks, supporting their role as versatile backbones. For baselines, we use LoRA’s original hyperparameters. In Solo Connection, we modify the rank and tune the learning rate. Our setup is consistent: one GPU, AdamW optimizer, batch size 4, and weight decay 0.1.

Table~\ref{tab:gpt1} reports BLEU, NIST, METEOR, ROUGE, and CIDEr scores \citep{hu2021lora, adalora}. Higher scores indicate better performance. {\color{blue}FT GPT-2 M}\footnote{Model variants are highlighted in blue for readability.} is the fully fine-tuned GPT-2 Medium with 354.92M parameters. Due to resource limits, we use previously reported results from \citep{hu2021lora}. We compare {\color{blue}Baseline LoRA GPT-2 M} (0.35M), {\color{blue}Solo Connection GPT-2 M} (0.26M), {\color{blue}Baseline LoRA GPT-2 S} (0.29M), and {\color{blue}Solo Connection GPT-2 S} (0.12M) in terms of trainable parameters.

Solo Connection consistently outperforms LoRA with fewer parameters. For GPT-2 Medium, it achieves a 99.93\% reduction over full fine-tuning, and 25.71\% over LoRA, with improved scores (+0.37\% BLEU, +0.70\% NIST, etc.). For GPT-2 Small, it cuts 58.62\% of parameters versus LoRA, while improving all metrics (+2.82\% BLEU, +1.76\% CIDEr, etc.). These results highlight Solo Connection’s efficiency in fine-tuning large language models for general-purpose tasks.

\section{Conclusion}

Large language models have demonstrated strong performance across NLP tasks, but adapting them to new domains remains computationally expensive—often inaccessible to labs with limited resources. To address this, we introduced \textit{Solo Connection}, a parameter-efficient fine-tuning method using long skip connections and decoder-block-level parameter sharing. Solo Connection achieves up to 59\% fewer trainable parameters than LoRA while consistently outperforming it on E2E generation benchmarks. Like LoRA, Solo Connection retains a compact set of adaptation weights, separate from the core model. These standalone modules can be independently swapped in and out, enabling multiple domain-specific adapters to run on a single GPU without duplicating the full backbone. This design significantly reduces resource overhead and enables faster, more cost-effective fine-tuning across diverse tasks.

\textbf{Future Work:} We aim to extend Solo Connection to broader datasets and model architectures—including computer vision—pending additional compute resources. This will help assess its scalability and impact across domains.

\bibliography{sample}

\begin{thebibliography}{30}
\providecommand{\natexlab}[1]{#1}
\providecommand{\url}[1]{\texttt{#1}}
\expandafter\ifx\csname urlstyle\endcsname\relax
  \providecommand{\doi}[1]{doi: #1}\else
  \providecommand{\doi}{doi: \begingroup \urlstyle{rm}\Url}\fi

\bibitem[Allgower \& Georg(2003)Allgower and Georg]{cont_method_Allgower}
Allgower, E. and Georg, K.
\newblock \emph{Introduction to Numerical Continuation Methods}.
\newblock Society for Industrial and Applied Mathematics, 2003.
\newblock \doi{10.1137/1.9780898719154}.
\newblock URL \url{https://epubs.siam.org/doi/abs/10.1137/1.9780898719154}.

\bibitem[Anonymous(2023)]{arxivpaper}
Anonymous.
\newblock The equivalence of finite and infinite impulse iterative neural networks.
\newblock \emph{in preparation for submission as an arXiv.org preprint}, 2023.

\bibitem[Ben~Zaken et~al.(2022)Ben~Zaken, Goldberg, and Ravfogel]{ben-zaken-etal-2022-bitfit}
Ben~Zaken, E., Goldberg, Y., and Ravfogel, S.
\newblock {B}it{F}it: Simple parameter-efficient fine-tuning for transformer-based masked language-models.
\newblock In \emph{Proceedings of the 60th Annual Meeting of the Association for Computational Linguistics (Volume 2: Short Papers)}, pp.\  1--9, Dublin, Ireland, May 2022. Association for Computational Linguistics.
\newblock \doi{10.18653/v1/2022.acl-short.1}.
\newblock URL \url{https://aclanthology.org/2022.acl-short.1}.

\bibitem[Dai et~al.(2019)Dai, Yang, Yang, Carbonell, Le, and Salakhutdinov]{txl}
Dai, Z., Yang, Z., Yang, Y., Carbonell, J., Le, Q.~V., and Salakhutdinov, R.
\newblock Transformer-xl: Attentive language models beyond a fixed-length context, 2019.

\bibitem[Doedel et~al.(2007)Doedel, Champneys, Dercole, Fairgrieve, Kuznetsov, Oldeman, Paffenroth, Sandstede, Wang, Zhang, et~al.]{doedel2007auto}
Doedel, E., Champneys, A., Dercole, F., Fairgrieve, T., Kuznetsov, Y.~A., Oldeman, B., Paffenroth, R., Sandstede, B., Wang, X., Zhang, C., et~al.
\newblock Auto-07p: Continuation and bifurcation software for ordinary differential equations.
\newblock 2007.

\bibitem[Gauthier et~al.(2021)Gauthier, Bollt, Griffith, and Barbosa]{reservoir}
Gauthier, D.~J., Bollt, E.~M., Griffith, A., and Barbosa, W. A.~S.
\newblock Next generation reservoir computing.
\newblock \emph{CoRR}, abs/2106.07688, 2021.
\newblock URL \url{https://arxiv.org/abs/2106.07688}.

\bibitem[He et~al.(2015)He, Zhang, Ren, and Sun]{HeZR015}
He, K., Zhang, X., Ren, S., and Sun, J.
\newblock Delving deep into rectifiers: Surpassing human-level performance on imagenet classification.
\newblock \emph{CoRR}, abs/1502.01852, 2015.
\newblock URL \url{http://arxiv.org/abs/1502.01852}.

\bibitem[He et~al.(2016)He, Zhang, Ren, and Sun]{resnet}
He, K., Zhang, X., Ren, S., and Sun, J.
\newblock Deep residual learning for image recognition.
\newblock In \emph{Proceedings of the IEEE Conference on Computer Vision and Pattern Recognition (CVPR)}, June 2016.

\bibitem[Hershey(2022)]{quincy}
Hershey, Q.
\newblock Exploring neural network structure through iterative neural networks: Connections to dynamical systems.
\newblock Master's thesis, Worcester Polytechnic Institute, 2022.

\bibitem[Hershey et~al.(2024)Hershey, Paffenroth, Pathak, and Tavener]{Pathak_rethink}
Hershey, Q., Paffenroth, R., Pathak, H., and Tavener, S.
\newblock Rethinking the relationship between recurrent and non-recurrent neural networks: A study in sparsity, 2024.
\newblock URL \url{https://arxiv.org/abs/2404.00880}.

\bibitem[Houlsby et~al.(2019{\natexlab{a}})Houlsby, Giurgiu, Jastrzebski, Morrone, De~Laroussilhe, Gesmundo, Attariyan, and Gelly]{adapter}
Houlsby, N., Giurgiu, A., Jastrzebski, S., Morrone, B., De~Laroussilhe, Q., Gesmundo, A., Attariyan, M., and Gelly, S.
\newblock Parameter-efficient transfer learning for {NLP}.
\newblock In Chaudhuri, K. and Salakhutdinov, R. (eds.), \emph{Proceedings of the 36th International Conference on Machine Learning}, volume~97 of \emph{Proceedings of Machine Learning Research}, pp.\  2790--2799. PMLR, 09--15 Jun 2019{\natexlab{a}}.
\newblock URL \url{https://proceedings.mlr.press/v97/houlsby19a.html}.

\bibitem[Houlsby et~al.(2019{\natexlab{b}})Houlsby, Giurgiu, Jastrzebski, Morrone, de~Laroussilhe, Gesmundo, Attariyan, and Gelly]{peft}
Houlsby, N., Giurgiu, A., Jastrzebski, S., Morrone, B., de~Laroussilhe, Q., Gesmundo, A., Attariyan, M., and Gelly, S.
\newblock Parameter-efficient transfer learning for {NLP}.
\newblock \emph{CoRR}, abs/1902.00751, 2019{\natexlab{b}}.
\newblock URL \url{http://arxiv.org/abs/1902.00751}.

\bibitem[Hu et~al.(2021)Hu, Shen, Wallis, Allen-Zhu, Li, Wang, Wang, and Chen]{hu2021lora}
Hu, E.~J., Shen, Y., Wallis, P., Allen-Zhu, Z., Li, Y., Wang, S., Wang, L., and Chen, W.
\newblock Lora: Low-rank adaptation of large language models, 2021.

\bibitem[Kopiczko et~al.(2024)Kopiczko, Blankevoort, and Asano]{vera}
Kopiczko, D.~J., Blankevoort, T., and Asano, Y.~M.
\newblock Ve{RA}: Vector-based random matrix adaptation.
\newblock In \emph{The Twelfth International Conference on Learning Representations}, 2024.
\newblock URL \url{https://openreview.net/forum?id=NjNfLdxr3A}.

\bibitem[LeCun et~al.(2015)LeCun, Bengio, and Hinton]{lecun2015deep}
LeCun, Y., Bengio, Y., and Hinton, G.
\newblock Deep learning.
\newblock \emph{nature}, 521\penalty0 (7553):\penalty0 436--444, 2015.

\bibitem[Nilesh~Pathak \& Paffenroth(2019)Nilesh~Pathak and Paffenroth]{pathak2018npacs}
Nilesh~Pathak, H. and Paffenroth, R.
\newblock Parameter continuation methods for the optimization of deep neural networks.
\newblock pp.\  1637--1643, 2019.
\newblock \doi{10.1109/ICMLA.2019.00268}.

\bibitem[Novikova et~al.(2017)Novikova, Du{\v{s}}ek, and Rieser]{novikova-e2e}
Novikova, J., Du{\v{s}}ek, O., and Rieser, V.
\newblock The {E}2{E} dataset: New challenges for end-to-end generation.
\newblock In Jokinen, K., Stede, M., DeVault, D., and Louis, A. (eds.), \emph{Proceedings of the 18th Annual {SIG}dial Meeting on Discourse and Dialogue}, pp.\  201--206, Saarbr{\"u}cken, Germany, August 2017. Association for Computational Linguistics.
\newblock \doi{10.18653/v1/W17-5525}.
\newblock URL \url{https://aclanthology.org/W17-5525}.

\bibitem[Pathak(2018)]{pathak2018parameter}
Pathak, H.~N.
\newblock \emph{Parameter continuation with secant approximation for deep neural networks}.
\newblock PhD thesis, Master's Thesis at Worcester Polytechnic Institute, 2018.

\bibitem[Pathak \& Paffenroth(2019)Pathak and Paffenroth]{pathak2019parameter}
Pathak, H.~N. and Paffenroth, R.
\newblock Parameter continuation methods for the optimization of deep neural networks.
\newblock In \emph{2019 18th IEEE International Conference on Machine Learning And Applications (ICMLA)}, pp.\  1637--1643. IEEE, 2019.

\bibitem[Pathak \& Paffenroth(2021)Pathak and Paffenroth]{pathakprincipled_harsh}
Pathak, H.~N. and Paffenroth, R.
\newblock Principled curriculum learning using parameter continuation methods.
\newblock 2021.

\bibitem[Pathak et~al.(2023)Pathak, Paffenroth, and Hershey]{PathakSeq2D}
Pathak, H.~N., Paffenroth, R., and Hershey, Q.
\newblock Sequentia12d: Organizing center of skip connections for transformers.
\newblock In \emph{2023 International Conference on Machine Learning and Applications (ICMLA)}, pp.\  362--368, 2023.
\newblock \doi{10.1109/ICMLA58977.2023.00057}.

\bibitem[Radford et~al.(2019)Radford, Wu, Child, Luan, Amodei, and Sutskever]{radford2019gpt2}
Radford, A., Wu, J., Child, R., Luan, D., Amodei, D., and Sutskever, I.
\newblock Language models are unsupervised multitask learners.
\newblock 2019.

\bibitem[Renduchintala et~al.(2024)Renduchintala, Konuk, and Kuchaiev]{renduchintala2024tiedlora}
Renduchintala, A., Konuk, T., and Kuchaiev, O.
\newblock Tied-lora: Enhancing parameter efficiency of lora with weight tying, 2024.

\bibitem[Ronneberger et~al.(2015)Ronneberger, Fischer, and Brox]{ronneberger2015unet}
Ronneberger, O., Fischer, P., and Brox, T.
\newblock U-net: Convolutional networks for biomedical image segmentation, 2015.

\bibitem[Rumelhart \& McClelland(1987)Rumelhart and McClelland]{rnn}
Rumelhart, D.~E. and McClelland, J.~L.
\newblock \emph{Learning Internal Representations by Error Propagation}, pp.\  318--362.
\newblock The MIT Press, 1987.

\bibitem[Sherstinsky(2020)]{Sherstinsky_2020_rnn}
Sherstinsky, A.
\newblock Fundamentals of recurrent neural network ({RNN}) and long short-term memory ({LSTM}) network.
\newblock \emph{Physica D: Nonlinear Phenomena}, 404:\penalty0 132306, mar 2020.
\newblock \doi{10.1016/j.physd.2019.132306}.

\bibitem[Strogatz()]{strogatz2001nonlinear}
Strogatz, S.~H.
\newblock \emph{Nonlinear dynamics and chaos: with applications to physics, biology, chemistry, and engineering (studies in nonlinearity)}, volume~1.

\bibitem[Touvron et~al.(2023)Touvron, Martin, Stone, Albert, Almahairi, Babaei, Bashlykov, Batra, Bhargava, Bhosale, Bikel, Blecher, Ferrer, Chen, Cucurull, Esiobu, Fernandes, Fu, Fu, Fuller, Gao, Goswami, Goyal, Hartshorn, Hosseini, Hou, Inan, Kardas, Kerkez, Khabsa, Kloumann, Korenev, Koura, Lachaux, Lavril, Lee, Liskovich, Lu, Mao, Martinet, Mihaylov, Mishra, Molybog, Nie, Poulton, Reizenstein, Rungta, Saladi, Schelten, Silva, Smith, Subramanian, Tan, Tang, Taylor, Williams, Kuan, Xu, Yan, Zarov, Zhang, Fan, Kambadur, Narang, Rodriguez, Stojnic, Edunov, and Scialom]{touvron2023llama2openfoundation}
Touvron, H., Martin, L., Stone, K., Albert, P., Almahairi, A., Babaei, Y., Bashlykov, N., Batra, S., Bhargava, P., Bhosale, S., Bikel, D., Blecher, L., Ferrer, C.~C., Chen, M., Cucurull, G., Esiobu, D., Fernandes, J., Fu, J., Fu, W., Fuller, B., Gao, C., Goswami, V., Goyal, N., Hartshorn, A., Hosseini, S., Hou, R., Inan, H., Kardas, M., Kerkez, V., Khabsa, M., Kloumann, I., Korenev, A., Koura, P.~S., Lachaux, M.-A., Lavril, T., Lee, J., Liskovich, D., Lu, Y., Mao, Y., Martinet, X., Mihaylov, T., Mishra, P., Molybog, I., Nie, Y., Poulton, A., Reizenstein, J., Rungta, R., Saladi, K., Schelten, A., Silva, R., Smith, E.~M., Subramanian, R., Tan, X.~E., Tang, B., Taylor, R., Williams, A., Kuan, J.~X., Xu, P., Yan, Z., Zarov, I., Zhang, Y., Fan, A., Kambadur, M., Narang, S., Rodriguez, A., Stojnic, R., Edunov, S., and Scialom, T.
\newblock Llama 2: Open foundation and fine-tuned chat models, 2023.
\newblock URL \url{https://arxiv.org/abs/2307.09288}.

\bibitem[Xu et~al.(2023)Xu, Xie, Qin, Tao, and Wang]{xu2023parameterefficient}
Xu, L., Xie, H., Qin, S.-Z.~J., Tao, X., and Wang, F.~L.
\newblock Parameter-efficient fine-tuning methods for pretrained language models: A critical review and assessment, 2023.

\bibitem[Zhang et~al.(2023)Zhang, Chen, Bukharin, Karampatziakis, He, Cheng, Chen, and Zhao]{adalora}
Zhang, Q., Chen, M., Bukharin, A., Karampatziakis, N., He, P., Cheng, Y., Chen, W., and Zhao, T.
\newblock Adalora: Adaptive budget allocation for parameter-efficient fine-tuning, 2023.

\end{thebibliography}
\bibliographystyle{icml2025}

\appendix
\section{Discussion and Related Work}

Recent advances in Parameter-Efficient Fine-Tuning (PEFT) have led to various techniques that effectively adapt pre-trained language models to specific tasks while minimizing additional parameters. Adapters \citet{adapter} are a prominent PEFT technique that has gained significant attention in recent years. The authors introduce additional learnable modules, called adapters, which are inserted between the layers of the pre-trained model. These adapters enable task-specific tuning while preserving the pre-trained knowledge. Adapters have demonstrated impressive performance gains in various NLP tasks, including language translation, sentiment analysis, and text classification. Adapters offer high flexibility and modularity, allowing easy integration with existing pre-trained models. 

The core motivation behind Solo Connection is to fundamentally shift the paradigm of parameter-efficient fine-tuning (PEFT) from intra-layer adaptation to inter-layer adaptation. While Solo Connection can be loosely described using the language of adapters, doing so overlooks its theoretical foundation and architectural intent. Most adapter-based methods, including LoRA and its variants, operate within layers—modifying specific submodules like attention or feedforward blocks.


LoRA \cite{hu2021lora} is another PEFT technique that adds low-rank matrices to the decoder layer's attention and feed-forward network layers. LoRA has demonstrated impressive performance gains in various NLP tasks and is widely used \citet{xu2023parameterefficient, hu2021lora}. Many methods have further advanced this technique to make this method more efficient and select appropriate rank \cite{peft, vera, adalora, xu2023parameterefficient}.  While Adapter \citet{adapter} and LoRA-based methods have impressive results, they all modify the internal workings of the decoder block. We refer to such PEFT techniques as intra-connections in GPT-2. In contrast, our proposed method, Solo Connection, operates outside the decoder layers, adapting the representation from one decoder block to another, as shown in Figure-\ref{fig:solor}. The Solo Connection approach offers a unique perspective on PEFT and has the potential further to improve the efficiency and effectiveness of PEFT techniques. 

 Skip connections, introduced in ResNet \cite{resnet}, allow deep neural networks to preserve original input information by bypassing certain layers. While widely used in various applications, their implementation in large language models like GPT-2 presents two significant challenges:

\textbf{Limited Scope in Current Models:} In GPT-2, skip connections are typically confined within individual decoder blocks. However, with models ranging from 12 to 128 or more decoder blocks, there's a pressing need to reevaluate the role and potential of skip connections during fine-tuning. This paper focuses on "long skip connections" that link outputs of different decoder blocks, potentially enhancing the model's ability to adapt to new tasks while leveraging pre-trained knowledge.

\textbf{Lack of Systematic Implementation:} Usually, skip connections have been applied ad-hoc, with architectures primarily driven by empirical results. This approach lacks a systematic framework for organizing and optimizing skip connections, especially in complex models. Recent work like Sequential2D \citet{PathakSeq2D, quincy, arxivpaper} has begun to address this by providing insights into organizing inter- and intra-connections in feedforward and decoder networks.

Our research builds upon these insights, particularly exploring the potential of "weighted long skip connections" for efficient GPT-2 fine-tuning. We draw inspiration from techniques like U-Net \citet{ronneberger2015unet}, which successfully employed long skip connections between convolutional blocks to achieve superior segmentation results.
By addressing these challenges, we aim to develop a more systematic and effective approach to implementing skip connections in large language models, potentially improving their adaptability and performance across various tasks.

Also, the idea of using shared blocks of training across neural networks is used in many works, for example, RNNs \citet{rnn, Sherstinsky_2020_rnn}, Tied-Lora \citet{renduchintala2024tiedlora} and VeRA \citet{vera} have used such technique to improve parameter efficiency. However, in this paper, we uniquely apply it for the decoder's representation learning and test this technique in Section-\ref{sec:experiment}. 


Parameter continuation methods \cite{doedel2007auto, pathak2019parameter, pathak2018npacs, cont_method_Allgower} are a way to adapt slowly from one continuous function to another. Parameter continuation methods and related numerical analysis techniques are widely used in Dynamical Systems, Bifurcations, and Chaos theory but have limited exposure to deep learning \citet{pathak2019parameter}. Model continuation methods \citet{pathak2019parameter, pathakprincipled_harsh}, where a simple neural network model is trained first, and gradually model is made complex, have demonstrated their effectiveness in achieving better training and generalization performance on specific unsupervised tasks.  In this paper, we mainly use the homotopy methods \citet{cont_method_Allgower, doedel2007auto, pathak2018npacs} to adapt the learning from the pre-trained phase to the fine-tuning phase gradually. While the homotopy method is easy to understand, the real challenge is where to apply it. In this paper, we discuss these details in Section-\ref{method}

We extend GPT-2 to adapt to newer tasks given the pre-trained knowledge continually. GPT-2 model is a transformer-based language model \citet{radford2019gpt2, PathakSeq2D} that consists of three main parts: the embedding layers, decoder blocks, and language model head. The Decoder block is the main computational block that we denote by $D_i$ that performs multi-headed attention and projection transformations to the input token vectors. We devise and utilize Solo Connection to enhance the decoder-block ($D_i$) representation for new tasks in a parameter-efficient way. Our proposed method, Solo Connection, uniquely adapts a different approach by connecting the output of one decoder block to the input of another and has components inspired and devised from roots of many well-established research works such as  Continuation methods, Skip Connection, and Fine-tuning of Deep learning models. Our approach offers a lightweight and efficient solution for pre-trained language model adaptation, making it an attractive alternative to LoRA and other PEFT techniques.

\section{Method: Calculation of Parameters for Fine-tuning}

This calculation determines the total number of parameters required for fine-tuning a pre-trained language model. Let us define some variables:
\begin{itemize}
    \itemsep0em
    \item $d$: the dimensionality of the decoder vector (1024 in this case)
    \item $r$: the dimensionality of the encoder vector (64 in this case)
    \item $s$: the sparsity factor (0.6 in this case, since $1 - s = 0.4$)
    \item $n$: the number of encoding and decoding units (2 in this case 1 for the encoder and 1 for the decoder)
    \item $T$: the number of decoder layers.
\end{itemize}
   
Here, we show step-by-step parameter count calculation. First, $d \cdot r \cdot n \cdot (1-s)$, which calculates the number of parameters required for the attention mechanism. It is the product of the decoder dimension, encoder dimension, number of attention heads, and the sparsity factor $(1 - s)$. Second term, $r \cdot T$, which calculates the number of parameters required for the encoder layers. It's the product of the encoder dimension and the number of decoder layers. Final term, $d \cdot T$ calculates the number of parameters required for the decoder layers. It's the product of the decoder dimension and the number of decoder layers.

The total number of parameters is the sum of these three components: $(d \cdot r \cdot n \cdot (1-s)) + r \cdot T + d \cdot T$.  For example, with $d = 1024$, $r = 32$, $s = 0.7$, $n = 2$, $T = 11$, $1 - s = 0.3$, the total number of parameters is 31,276. So, the total number of parameters required for fine-tuning in this case it would be 31,276. Note that these calculations are specific to the paper's architecture and may vary depending on the model and fine-tuning setup.



\section{More Experiments}\label{a:exp}
\begin{table*}[h]
\centering
\begin{tabular}{|p{6cm}|c|c|c|c|c|c|}
\hline
Model & \#Params & BLEU & NIST & E2E MET & Rouge & CIDEr \\
\hline
  Baseline LoRA GPT-S & 0.29M & 65.79 & 8.49 & 45.21 & 67.46 & 2.27 \\
\hline
 Solo Connection (r=512) & 0.8M & \cellcolor{green!80}67.28 & \cellcolor{green!80}8.60 & \cellcolor{green!80}45.93 & \cellcolor{green!80}68.09 & \cellcolor{green!80}2.33 \\
Solo Connection (r=128) & 0.2M & \cellcolor{green!80}66.39 & \cellcolor{green!80}8.52 & 44.95 & \cellcolor{green!80}68.07 & \cellcolor{green!80}2.27 \\
Solo Connection (r=64) & 0.14M & 65.57 & \cellcolor{green!80}8.58 & 43.66 & 65.99 & 2.20 \\
Solo Connection (r=32) & 0.078M & 65.30 & 8.53 & 43.35 & 65.30 & 2.17 \\
 Solo Connection (r=8) & 0.02M & 63.93 & 8.04 & 40.15 & 65.27 & 1.97 \\

\hline
 Solo Connection (r=512, s=0.6) & 0.47M & \cellcolor{green!80} 67.64 & \cellcolor{green!80}8.65 & \cellcolor{green!80} 45.70 & \cellcolor{green!80} 68.28 & \cellcolor{green!80} 2.33 \\
Solo Connection (r=128, s=0.6) & 0.12M & \cellcolor{green!80} 67.64 & \cellcolor{green!80}8.64 & \cellcolor{green!80} 45.70 & \cellcolor{green!80} 68.32 & \cellcolor{green!80} 2.31 \\
 Solo Connection (r=64, s=0.6) & 0.06M & \cellcolor{green!80} 67.72 & \cellcolor{green!80}8.68 & 45.09 & 67.18 & 2.30 \cellcolor{green!80} \\
 Solo Connection (r=32, s=0.6) & 0.03M & \cellcolor{green!80} 67.50 & \cellcolor{green!80} 8.6 & 45.46 \cellcolor{green!80} & \cellcolor{green!80} 68.38 & 2.31 \cellcolor{green!80}\\
 Solo Connection (r=8, s=0.6) & 0.011M & 65.84 & 8.46 & 43.06 & 66.40 & 2.18 \\

\hline
\end{tabular}
\caption{Comparison of performance metrics for various GPT-2 Small fine-tuned using different methods. The table displays the number of parameters (Params) for each model. Additionally, BLEU, NIST, E2E MET, Rouge, and CIDEr scores are provided for each model, facilitating comparison of the performance across different fine-tuning methods. }
\label{tab:gpt_s}
\end{table*}

\subsection{What is the impact of rank and sparsity?}

\begin{table*}[h]
\centering
\begin{tabular}{|p{6cm}|c|c|c|c|c|c|}
\hline
Model & \#Params & BLEU & NIST & E2E MET & Rouge & CIDEr \\
GPT-2 M (FT)* & 354.92M & 68.2 & 8.62 & 46.2 & 71.0 & 2.47 \\
\hline
(LoRA) Baseline & 0.35M & 67.45 & 8.58 & 45.93 & 68.8 & 2.36 \\
\hline
Solo Connection (r=512) & 1.06M & \cellcolor{green!80}68.10& \cellcolor{green!80} 8.65 & 45.32 & 68.13 & 2.33 \\ 
Solo Connection (r=128) & 0.27M & \cellcolor{green!80}67.89 & \cellcolor{green!80} 8.66 & 45.29 & 68.56 & \cellcolor{green!80} 2.34 \\
Solo Connection (r=64) & 0.14M & 65.30 & 8.58 \cellcolor{green!80} & 42.96 & 63.52 & 2.18 \\
Solo Connection (r=32) & 0.077M & 64.93 & 8.54 & 43.15 & 63.27 & 2.17 \\
Solo Connection (r=8) & 0.027M & 64.93 & 8.54 & 43.15 & 63.27 & 2.17 \\
\hline
Solo Connection (r=512, s=0.7) & 0.26M & \cellcolor{green!80} 67.7 & \cellcolor{green!80} 8.64 & \cellcolor{green!80}45.95 & \cellcolor{green!80} 69.13 & \cellcolor{green!80}2.36\\
Solo Connection (r=128, s=0.7) & 0.09M & 67.04 & \cellcolor{green!80} 8.58 & 45.85 & 68.55 & 2.31\\
Solo Connection (r=64, s=0.6) & 0.06M & 65.72 & 8.45 & 45.27 & \cellcolor{green!80} 69.0 & 2.30\\
Solo Connection (r=32, s=0.6) & 0.037M & 66.36 & 8.5 & 44.71 & 67.6 & 2.32\\
Solo Connection (r=8, s=0.7) & 0.014M & 63.06 & 8.22 & 42.56 & 64.6 & 2.11\\
\hline

\hline
\end{tabular}
\caption{Performance metrics for GPT2-M (medium) models. Comparison of performance metrics for various GPT-based models fine-tuned using different methods. The table displays the number of parameters (Params) for each model. Additionally, BLEU, NIST, E2E MET, Rouge, and CIDEr scores are provided for each model, facilitating comparison of the performance across different fine-tuning methods. }
\label{tab:gpt_m}
\end{table*}

 In Table-\ref{tab:gpt_s} and Table-\ref{tab:gpt_m}, we show comparison between the various values of the two most important hyperparameters of the Solo Connection i.e. Low-rank dimension and sparsity. 
 
 To achieve this, we ran a set of experiments with variable ranks such as 8, 32, 64, 128, and 512 and similarly for the sparsity parameter 0.6, and 0.7. For comparison purposes, we made two groups with and without sparsity to see the clear difference in performance and listed results for GPT-2 M in Table-\ref{tab:gpt_m} and GPT-2 S in Table-\ref{tab:gpt_s}. In Table-\ref{tab:gpt_s}, we saw most Solo Connection with sparsity performed better than the Solo Connection without sparsity across all hyperparameter settings.

\subsection{Impact of Number of Skip Connections and Solo Connections on Performance}
We now evaluate the impact of varying both the number of Solo Connections and their span across the decoder. In the previous section, we applied a single Solo Connection to adapt one decoder block at a time. Here, we explore configurations where multiple Solo Connections are used, each spanning a longer range of decoder blocks. Specifically, we experiment with setups where three and five consecutive decoder blocks share a single Solo Connection.
These experiments remain highly efficient, as they require minimal trainable parameters. Table \ref{tab:zoom_factor_solo} presents the performance metrics for different configurations, analyzing both the number of Solo Connections and their span across the decoder.
Our results indicate that increasing the Solo Connection span to three blocks maintains performance close to our best results in Table-\ref{tab:gpt1}. However, when the span is extended to five blocks, we observe a significant performance drop, especially when the overall number of Solo Connections is reduced. These trends remain consistent across different Solo Encoder dimension settings (i.e., 64 and 128).

\begin{table*}[!h]
\centering
\begin{tabular}{|p{6cm}|c|c|c|c|c|c|}
\hline
Model & \#Params & BLEU & NIST & E2E MET & Rouge & CIDEr \\
\hline
Solo Connection (128, span=3, s=0.6) & 78k  & 65.73 & 8.50 & 44.52 & 67.14 & 2.23 \\
Solo Connection (64, span=3, s=0.6) & 41k & 66.67 & 8.54 & 44.35 & 66.45 & 2.21 \\
Solo Connection (128, span=5, s=0.6) & 76k  & 28.75 & 2.54 & 18.44 & 30.95 & 0.31 \\
Solo Connection (64, span=5, s=0.6) & 38k & 27.28 & 2.90 & 17.93 & 30.06 & 0.33 \\
\hline
\end{tabular}
\caption{Performance metrics evaluating the impact of increasing the Solo Connection span to cover multiple decoder blocks. The table highlights how varying the span of Solo Connections affects model performance. Results for span=1 are in the table-\ref{tab:gpt_s} and is the top performer followed by  3 and 5. Span=3 also shows promising results with fewer trainable parameters than span=1}
\label{tab:zoom_factor_solo}
\end{table*}

\subsection{Should the matrices, Encoder and Decoder be trained?}

Next, after doing an extensive literature survey, we found that the efficiency of LoRA can be further improved using methods such as sparsity, adaptive rank, and parameter sharing \citet{hu2021lora, adalora, vera}. We test the hypothesis of whether the individual components encoder and decoder should be trained or the random and non-trainable transformations can yield similar results as observed in recent literature \citet{reservoir, vera}. In Table-\ref{tab:random_m}, we show results with two ranks (r=512 and r=1024), and with both, the generalization results on all the metrics were poor for the case of random and non-trainable functions. 

\begin{table*}[!h]
\centering
\begin{tabular}{|p{6cm}|c|c|c|c|c|c|}
\hline
Model & \#Params & BLEU & NIST & E2E MET & Rouge & CIDEr \\
\hline
Solo Connection (r=512) & 0.8M & 67.28 & 8.60 & 45.93 & 68.09 & 2.33 \\
Solo Connection (r=512) & 9k & 53.77 & 4.8069 & 34.89 & 61.25 & 1.38 \\ 
Solo Connection (r=1024) & 10k & 61.56 & 7.1842 & 38.57 & 64.56 & 1.7116 \\

\hline
\end{tabular}
\caption{Performance metrics where Solo connection's encoder and decoder are random and not trainable.}
\label{tab:random_m}
\end{table*}

\subsection{Affect of Homotopy Linear Layer}

In this section, we examine the importance of the homotopy parameter when fine-tuning GPT-2 Small with Solo Connection. We conduct two experiments - Case-1: Solo Connection includes a homotopy layer, as defined in Equation~\ref{eq:homotopy}. Case-2: The homotopy layer is replaced with a simple trainable vector $g(\textbf{z}) =  \textbf{v} \odot \textbf{z}$ where $\textbf{z}$ is the input vector and $\textbf{v}$ is the output. 

Table~\ref{tab:homo} highlights the critical role of the homotopy layer in training Solo Connection. Specifically, we find that removing the homotopy parameter $\lambda$ leads to a training collapse from which the model does not recover, even after multiple epochs. A closer investigation reveals that, in Case-1, the final value of $\lambda$ post-fine-tuning typically falls between 0 and 0.1, thereby normalizing the contribution from the Solo Connection. In contrast, in Case-2, the random initialization of the trainable vector hinders convergence relative to Case-1. In future work, we plan to conduct additional experiments to further explore this configuration.

\subsection*{Limitations}
We acknowledge that recent PEFT variants offer promising advancements, such as SVF, SVFT, MiLoRA, PiSSA, LoRA‑XS, and ProLoRA. However, many of these methods have not yet reported evaluation results on the E2E benchmark, which is central to our study. Moreover, these approaches predominantly focus on intra-layer adaptations (modifying weight matrices within individual transformer blocks), whereas Solo Connection introduces trainable inter-layer skip connections. This distinct architectural choice and a novel homotopy-based adaptation mechanism enable efficient cross-layer representation sharing, setting our approach apart. We commit to incorporating broader baseline comparisons in future work once resources permit.

\begin{table*}[!h]
\centering
\begin{tabular}{|p{6cm}|c|c|c|c|c|c|}
\hline
Model & \#Params & BLEU & NIST & E2E MET & Rouge & CIDEr \\
\hline
Homotopy Layer (Case-1)  & 0.12M &  67.64 & 8.64 &  45.70 &  68.32 &  2.31 \\
\hline
Linear Vector (Case-2) & 0.12M &  0.0 & 0.89 & 0.02 &  0.13 &  0.002 \\

\hline
\end{tabular}
\caption{Comparing performance metrics for Solo connection (r=128, s=0.6) with trainable Homotopy Layer (Case-1) to the performance with a simple trainable vector (Case-2).}
\label{tab:homo}
\end{table*}

\section{Impact Statement}

This paper seeks to advance the field of Machine Learning by introducing a more efficient approach to training Large Language Models (LLMs). By reducing the computational and resource demands associated with fine-tuning, our method has the potential to make advanced language modeling accessible to a broader range of researchers, industries, and organizations. Many people can benefit from this work, for instance by incorporating sophisticated natural language generation or understanding features into applications without incurring prohibitive costs.

Our work has many potential societal implications—both beneficial and unintended. While we do not identify any particular risks that warrant specific emphasis here, we acknowledge that making LLMs easier to develop and deploy could have downstream effects on misinformation, privacy, and bias. We encourage practitioners and researchers to consider these broader impacts when applying our method thoughtfully and to adopt responsible deployment practices to ensure equitable and ethical use of large-scale language models.

\end{document}